\documentclass{article}
\usepackage{ijcai23}

\usepackage{times}
\usepackage{soul}
\usepackage{url}
\usepackage[hidelinks]{hyperref}
\usepackage[utf8]{inputenc}
\usepackage[small]{caption}
\usepackage{graphicx}
\usepackage{amsmath}
\usepackage{amsthm}
\usepackage{amsfonts}
\usepackage{booktabs}
\usepackage{algorithm}
\usepackage[noend]{algorithmic}
\usepackage[switch]{lineno}
\usepackage{xcolor} 
\usepackage{amsmath}
\usepackage{tikz}
\usepackage{amsthm}
\usepackage{overpic}
\usepackage{natbib}
\usepackage{comment}

\newtheorem{definition}{Definition}
\newtheorem{lemma}{Lemma}
\newtheorem{proposition}{Proposition}
\newtheorem{remark}{Remark}

\newtheorem{theorem}{Theorem}

\title{Safe Multi-agent Learning via Trapping Regions}

\author{
Aleksander Czechowski
\and
Frans A. Oliehoek
\affiliations
Delft University of Technology
}

\begin{document}

\maketitle

\let\thefootnote\relax\footnotetext{Due to space restrictions, the proofs of our theorems and lemmas have been postponed to the Supplementary Material, and can be accessed in the extended version of the paper: \cite{arXivPaper}.
There, we also provide an additional example, where we find trapping regions in a model of economic competition, which ensures that none of the competing companies will reduce their production to zero.}

\begin{abstract}
One of the main challenges of multi-agent learning lies in establishing convergence of the algorithms, as, in general, a collection of individual, self-serving agents is not guaranteed to converge with their joint policy, when learning concurrently. This is in stark contrast to most single-agent environments, and sets a prohibitive barrier for deployment in
 practical applications, as it induces uncertainty in long term behavior of the system.
 In this work, we apply the concept of trapping regions, known from qualitative theory of dynamical systems, to create safety sets in the joint strategy space for decentralized learning.
 We propose a binary partitioning algorithm for verification that candidate sets form trapping regions in systems with known learning dynamics, and a heuristic sampling algorithm for scenarios where learning dynamics are not known.
 We demonstrate the applications to a regularized version of Dirac Generative Adversarial Network,  a four-intersection traffic control scenario run in a state of the art open-source microscopic traffic simulator SUMO, and a mathematical model of economic competition.
\end{abstract}

\section{Introduction}
In the recent years, enormous progress has been made
for single agent planning and learning algorithms, with agents matching or exceeding human performance 
in various tasks and games~\citep{mnih2013playing,silver2016mastering}.
The vast success of single agent learning can be partially explained by robustness and strong convergence properties of the underlying algorithms in their basic form, such as Q-learning~\citep{watkins1992q} or policy gradients~\citep{sutton1999policy}. 
Despite wide interest, the same cannot be however said for multi-agent learning.
Even most basic models, e.g. replicator learning for normal form games, exhibit nonconvergence, cyclic or even chaotic behavior~\citep{Sato2}.
Even worse, it has been shown that in decoupled learning systems,
there can be no learning rule that guarantees convergence to a Nash equilibrium~\citep{Hart}. 
These nonconvergent examples have also been found in more practical learning problems, such as Generative Adversarial Networks~\citep{mescheder2018training}.
Due to the above limitations, 
many successful multi-agent learning
methods resorted to using a centralized component, such as centralized critic in actor-critic learning~\citep{lowe2017multi} or 
meta-solvers in double-oracle type algorithms like Parallel Nash Memory~\citep{oliehoek2006parallel} and Policy Space Response Oracles~\citep{lanctot2017unified}.
Alternatives in form of decentralized algorithms usually rely on specific assumptions on the reward structure for convergence, 
for instance fictitious play or exploitability descent in zero-sum games~\citep{brown1951iterative,exploit}.

Despite this undisputed progress in designing convergent multi-agent algorithms, it can be argued that in practical, real-world multi-agent scenarios there will be plenty of situations, where convergence cannot be enforced. One can easily envisage a situation, e.g. in automated traffic control and driving scenarios, or in automated trading, where multiple entities (be it traffic lights, vehicles, or brokers) follow own learning protocols for individual reward maximization. Such learning rules, even though designed to be convergent in static, single-agent environments, would invariably interfere with one another in a multi-agent setting, sometimes resulting in cyclic, or divergent outcomes.
The lack of convergence guarantees in such general settings forms a major obstacle for introduction of online learning systems in practical applications, as it introduces a lot of uncertainty over what will be the state of the system, if learning is left unsupervised.
Can we nevertheless still establish a type of safety certificates, that would allow us to conclude that simultaneous learning will not spin out of control?
%%%%%%%%%%%%

In this paper, we suggest a novel approach to address issue. We start from the realization, that convergence is often not absolutely necessary for reliability. From systems designers perspective, it is often enough to know that learning has \emph{rough} stability guarantees -- that is, that agents will not leave a predetermined region of the strategy space during learning.
For a conceptual application, let us consider a traffic light control network as in Figure~\ref{fig:traf}, where individual traffic light controllers learn the best balance of green time between the phases to minimize the waiting time for approaching vehicles.
%as depicted in Figure~\ref{fig0}.
It is not absolutely necessary that by learning each intersection reaches a final, static setting, but would be essential that at all times it gives minimal green time of at least few seconds to each phase, to make sure that no vehicle gets stuck indefinitely on a red traffic light, and also serve the lingering pedestrian flows.

We propose a method of \emph{a priori} verifying
these constraints, by establishing \emph{trapping regions}; regions of strategy space, which learning trajectories will never escape.
The idea behind this concept is simple: a candidate set for a trapping region is formed by the constraints imposed by practical, problem-dependent safety considerations. By verifying whether such set is forward-invariant for the joint learning operator, we obtain a yes--or--no answer on whether it is safe to allow multi-agent learning (possibly in a decentralized manner), without  breaking these  constraints.  
This method can be seen an alternative solution concept in systems, where Nash equilibria are difficult or impossible to reach by learning dynamics. 
Trapping regions are intended to be used as a safety prerequisite. For instance a road authority could pre-approve the algorithms of automated road users, by checking whether their joint policy forms suitable trapping regions -- before they are deployed in real life.

This paper is organized as follows. Section 2 introduces the setting and necessary preliminaries. In Section 3 we present the definition of a trapping region, prove several useful theorems and lemmas that are useful for their verification for Lipschitz-continuous learning, and present two algorithms, for verifying whether given hyperrectangular sets forms a trapping region, only from the knowledge of learning operator on the set boundary. The first algorithm, is based on binary partitioning, and is applicable when learning dynamics are known analytically and Lipschitz, and we would like to have a mathematically rigorous guarantee. The second one is a heuristic algorithm, applicable in scenarios where learning dynamics can only be sampled, its dependability is directly correlated to the number of boundary samples taken.

Finally, in Section 4 we introduce two examples, that illustrate the applications of trapping regions.
The first of our first of our examples is a toy problem, a simple GAN-like learning scenario, where gradient learning starting from almost all points never converges, whereas trapping regions are abundant and easy to find. In our second example we deal with a practical traffic control problem, where four intersection in a traffic network adjust their strategies to dispatch traffic in an optimal manner. For this problem, we construct and verify a trapping region, which ensures all traffic directions will be given enough time, when traffic controllers are left to learn unsupervised.

\subsection{Related work} 

Trapping regions are well known and standard tool in qualitative theory of dynamical systems, e.g.~\citep{meiss2007differential,bonatti2006generic},
but to the best of our knowledge have not been directly applied 
in learning and control scenarios.
The majority of work on safety guarantees in control theory focuses on so-called constrained optimization~\citep{altman1999constrained}.
In the context of safe reinforcement learning, the focus has been on designing algorithms that 
satisfy particular safety constraints, c.f.
~\citep{garcia2015comprehensive} and references therein.
In the multi-agent case, research has been directed towards methods where an orchestrator~\citep{elsayed2021safe} or agents individually~\citep{lu2021decentralized} 
are adapting their behavior to respect the constraints; 
this has also been the underlying philosophy in the method of \emph{barrier functions} ~\citep{barrier1,barrier2}. There are also strong connections to methods of formal verification methods, in particular ones based on reachability analysis~\citep{ruan2018reachability, wang2021formal}.

\paragraph{Relation to Lyapunov control } 
Our method shares most similarities with the ones based on Lyapunov functions, such as Neural Lyapunov Control~\cite{Lyap}, see also~\cite{Berkenkamp}.
There are however several key differences that we would like to highlight here.
Most importantly, Lyapunov-based methods are only applicable to
(locally) convergent scenarios, as the existence of a Lyapunov function implies the existence of a locally attracting equilibrium of the system.
On the other hand, trapping regions are very well suited to deal with problems, where learning never converges to a stationary solution. Even if the learning trajectory does not converge, the bounds provided by the trapping region will ensure that they never diverge into unsafe regions of the joint strategy space. This has strong practical consequences: a non-convergent learning process without safety guarantees would have to be indefinitely supervised, whereas trapping regions ensure that it only explores safe parts of the strategy space, and does not require supervision.
Dealing with non-stationarity is particularly important for multi-agent systems, as it was shown e.g. in~\cite{kleinberg} outcomes of a non-convergent cyclic learning process can lead to higher social welfare than these of a stationary Nash; and even worse, often stationarity cannot be ensured, as some learners can be outside of our control (e.g. in adversarial scenarios).
In Section~\ref{sec:gan}, we provide a low-dimensional example where none of the learning trajectories converge to the equilibrium point, but trapping regions are easy to find, c.f. Figure~\ref{fig:my_label}.

The second difference comes in computational complexity. Neural Lyapunov Control requires evaluation of learning directions over a whole domain, while we only require it on a boundary of a domain, effectively reducing the dimension of the verification problem by one (c.f. Lemma~\ref{lem1}).

\section{Preliminaries}

We consider decentralized learning schemes for groups of $n$ agents that can be represented compactly by discrete adaptive dynamics of the form:
\begin{align}\label{eq1}
\begin{aligned}
    x^1_{t+1} &:= x^1_t + \gamma F_1(x^1_t,\dots , x^n_t),\\
    &\dots,\\
    x^n_{t+1} &:= x^n_t + \gamma F_n(x^1_t,\dots , x^n_t),
        \end{aligned}
\end{align}
where $x_i \in X_i \subset \mathbb{R}^{k_i}$ represents a point in the strategy space of a given agent $i$ (e.g. weights in a neural network or ratios of playing a mixed strategy), and the parameter $\gamma \in \mathbb{R}^+$ denotes the adaptation rate. Throughout this paper, we assume that the learning operators are continuous, and we denote by $N=\sum_i k_i$ the dimensionality of the joint learning space.
The maps $F_i: X_i \to \mathbb{R}^{k_i}$ represent the \emph{learning operators}, i.e. the outputs of the algorithms of each agent based on the inputs. For instance, for individual gradient-ascent type of algorithms we have 
\begin{align}\label{eq2}
    \begin{aligned}
    F_i(x^1, \dots, x^n) &= \nabla_{x_i} \mathbb{E}(R_i | x^1, \dots, x^n).
    \end{aligned}
\end{align} 
with $R_i: \mathbb{R}^N \to \mathbb{R}$ being the individual reward/payoff for agent $i$. To simplify the exposition, we will sometimes represent the learning system~\eqref{eq1} in a vectorized notation 

\begin{equation}\label{eq3}
    x_{t+1}: =x_{t}+\gamma F(x_t)
\end{equation}
with $F=[F_1, \dots, F_n]^T$ and $x=[x^1, \dots , x^n]^T$. Joint strategy sequences $\{x_t\}_t$ which satisfy~\eqref{eq3} will be referred to as the \emph{learning trajectories}.

An \emph{equilibrium} for the system~\eqref{eq3} is a point in the joint strategy space $x_* \in \mathbb{R}^N$ such that $F(x_*)=0$. In gradient learning, it is also a necessary condition for a strategy profile to be a \emph{local optimum}, with the sufficient condition being that the Hessian of the learning operator $F$ is negative definite. We remark that an equilibrium for the learning system~\eqref{eq3} does not necessarily need to be a Nash equilibrium; however, a local optimum is a \emph{local Nash point}, i.e. no agent is able to increase their reward unilaterally from a such point by performing a small deviation in its strategy.

For single agent learning, gradient descent in~\eqref{eq2} does converge to a local optimum under mild assumptions of regularity of $R_i$, and suitable choices of $\gamma$ (i.e. $\gamma$ can be constant, but needs to be suitably small). 
In general multi-agent setting, learning schemes given by systems of form~\eqref{eq1} can have complicated, even chaotic dynamics, and might not converge to equilibria at all, as for instance in relatively simple two-player games~\citep{Sato2}.

\section{Trapping regions}
Convergence in multi-agent learning cannot be always guaranteed; however the key aspect for security / reliability is often enough to ensure, that learning agents do not diverge into regions of policy space, which can yield dangerous combinations of strategies.
To this end, one needs to contain the learning trajectories within a prescribed safety region. Motivated by this rationale, in this section we formally define the \emph{trapping region} -- a subset of the joint strategy space, characterized by the property that learning trajectories that begin within such region can never leave it. 

We remark that the definitions and theorems below could have been framed in a continuous learning
setting by working with the ordinary differential equation 
$    \dot{x}=F(x) $.
but we opted for a discrete point of view, as more commonly encountered in literature on learning systems. 
The discrete system~\eqref{eq3} does in fact emerge as the Euler numerical solution of the ODE, with step size $\gamma$.

\subsection{Formal definition and forward invariance}
In what follows, we will denote by $\operatorname{int} X$ and $\partial X$ respectively
the topological interior and boundary of a set $X$, by $\operatorname{dist}(x,X)$ the Euclidean distance between a point $x$ and a set $X$, and by $\operatorname{diam}(X)$ the diameter (in Euclidean distance) of a set $X$. By $X^l$ we will denote the Cartesian product of $l$ copies of $X$.
%We will also use the Hausdorff distance between sets,
%defined by
%$\operatorname{dist}_{H}(X,Y)=\max\{\sup\{\operatorname{dist}(x,Y):x\in X\}, \{\operatorname{dist}(y,X), y \in Y\} \}$.
%
We also recall that a compact set is a set which is bounded, and
which contains limit points for all convergent sequences of its elements. We first recall the classical definition of a trapping region in context of learning dynamics~\eqref{eq1}.

\begin{definition} c.f. \citep{bonatti2006generic}.
Let $\mathbf{T} \subset \mathbb{R}^N$ be a compact subset of the joint strategy space, and let $\gamma > 0$.
If \begin{equation}\label{defeq}
x + \gamma F(x) \subset \operatorname{int} \mathbf{T},\quad \forall x \in \mathbf{T},\end{equation} 
then we call $\mathbf{T}$ a trapping region (for the system~\eqref{eq3}, with learning rate $\gamma$). 
\end{definition}
The following theorem, a folklore in dynamical systems community, highlights the advantage of establishing trapping regions; a trapping region not only guarantees that the learning curves starting inside can never escape it, but also that there is a learning equilibrium (possibly a Nash equilibrium) inside of it. 

\begin{theorem}\label{thm1}
 Let $\mathbf{T}$ be a trapping region. Then
\begin{enumerate}
    \item Any learning trajectory~\eqref{eq3} that starts in $\mathbf{T}$ never leaves $\mathbf{T}$,
    \item If $\mathbf{T}$ is convex, then there exists a learning equilibrium $x^* \in \operatorname{int} \mathbf{T}$.
\end{enumerate}
\end{theorem}

\subsection{Algorithmic verification: explicit learning dynamics}

In practice, verification of condition~\eqref{defeq} can be troublesome, as the volume of the trapping region usually requires a prohibitively high amount of samples.
For small learning rates and continuous learning dynamics, it is however enough to verify this assumption on the boundary, as any trajectory that could leave the region would have to pass through the boundary area. This is a standard argument, more commonly known in the continous case (e.g.~\citep{meiss2007differential}), and is formalized for our discrete setting in Lemma~\ref{lem1}.

\begin{lemma}\label{lem1}
Given a compact set $\mathbf{T}$, if $\gamma>0$ is sufficiently small, and for all $x \in \partial \mathbf{T}$ we have
\begin{equation}\label{bdtrap}
x + \gamma F(x) \in \operatorname{int} \mathbf{T},\end{equation} then $\mathbf{T}$ is a trapping region.
\end{lemma}

% it is enough to split to boundary and interior; then analyze the interior case and construct a sequence of maps; counterexample for bigger gamma annulus boundaries go to median interior goes to interior

Lemma~\ref{lem1} can be used to derive
exact inequalities needed to be satisfied by the learning operators, 
which are sufficient to establish a trapping region. 
In what follows, we denote by $x^i=[x^{i1}, \dots , x^{ik_i}]^T$ and $F_i=[F_{i1}, \dots , F_{ik_i}]^T$ the components of strategies and learning operators for each agent $i \in 1,\dots ,n$. In the examples we will sometimes omit the second subscript, when the strategy space of each agent is one-dimensional.

\begin{definition}
Let $\mathbf{T}$ be a set of the form of a product of intervals
\begin{equation}\mathbf{T}:=[x^{11}_-,x^{11}_+] \times \dots \times [x^{nk_n}_-,x^{nk_n}_+] \subset \mathbb{R}^N.\end{equation}
For $i \in 1,\dots,n$, $j \in 1,\dots k_i$, 
we denote by $\mathbf{T}_l^{ij}$ the set of all points $x \in \mathbf{T}$,
such that $\pi_{ij}$ -- the projection onto $i$-th agents $j$-th component satisfies $\pi_{ij} x = x_-^{ij}$.
We call this set \emph{the ($ij$th) left face} of $\mathbf{T}$.
Similarly, we denote by $\mathbf{T}_r^{ij}$ the set of all points $x \in \mathbf{T}$,
such that $\pi_{ij} x = x_+^{ij}$,
and call it \emph{the ($ij$th) right face} of $\mathbf{T}$.
\end{definition}

Our next Lemma follows follows directly from Lemma~\ref{lem1} applied to a trapping region of form of a product of intervals.

\begin{lemma}\label{lem2}

Given a set $\mathbf{T} \in \mathbf{R}^N$ which is a product of intervals,
assume that the following \emph{ isolation inequalities} are satisfied:
\begin{align}\label{isolation}
    \begin{aligned}
        F_{ij}(x) &> 0, \ \forall x \in \mathbf{T}_l^{ij},\\
        F_{ij}(x) &< 0, \ \forall x \in \mathbf{T}_r^{ij}.
        %F_i\left( \prod_{l=1}^{i-1} \prod_{m=1}^{k_l} [x_{lm}^-, x_{lm}^+]  \times x_{ij}^- \times \prod_{l=i+1}^{n} \prod_{m=1}^{k_l} [x_{lm}^-,x_{lm}^+] \right) &> 0,\\
        % F_i\left( \prod_{k=1}^{i-1} [x_k^-, x_k^+]  \times x_i^+ \times \prod_{k=i+1}^N [x_k^-,x_k^+] \right) &< 0.
    \end{aligned}
\end{align}
Then, the set $\mathbf{T}$ is a trapping region for $\gamma>0$ sufficiently small.
\end{lemma}

For Lipschitz-continuous learning dynamics, and trapping regions of form of a product of intervals, explicit bound on the range of $\gamma$ can be given.

\begin{theorem}\label{thm2}
Let $\mathbf{T}$ be as in Lemma~\ref{lem2} and $F$ be Lipschitz-continuous with Lipschitz constant over $\mathbf{T}$ bounded from above by $L$. The upper bound on step size $\gamma$ for which $\mathbf{T}$ forms a trapping region in the learning system~\eqref{eq1} can be given explicitly by
\begin{equation}\label{bound}
\gamma < \frac{ \min_{p\in\{l,r\}}\min_{ij} \min_{x \in \mathbf{T}^{ij}_{p}}|F_{ij}(x)|}{L\max_{x \in \mathbf{T}}{||F(x)||_{\text{max}}}}.
\end{equation}
\end{theorem}

\begin{remark}
For sufficiently regular boundaries $\partial \mathbf{T}$, conditions~\eqref{isolation} can be generalized to situations, where $\mathbf{T}$ is not a product of intervals. Namely, for $\mathbf{T}$ to be a trapping region it is enough that
\begin{equation}
 \langle F(x), n_{\partial \mathbf{T}}(x) \rangle < 0, \forall x \in \partial \mathbf{T},\end{equation} where $n_{\partial \mathbf{T}}(x)$ is the normal vector to $\partial \mathbf{T}$, pointing in direction outwards of $\mathbf{T}$, c.f.~\cite{meiss2007differential}.
\end{remark}

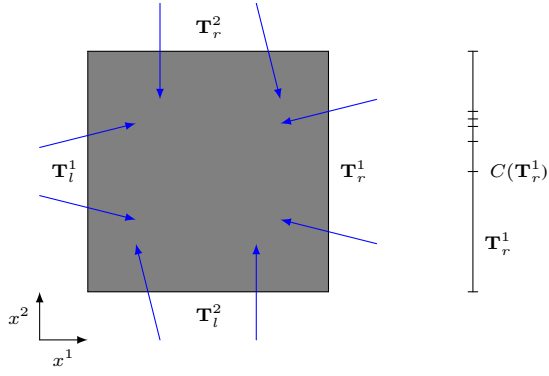
\begin{figure}
\begin{tikzpicture}[line cap=round,line join=round,>=latex,x=2mm,y=2mm,scale=1.6]
 
  \clip(-12,-8.4) rectangle (19,7);

  \fill [color=gray] (-5.,-5.) -- (-5.,5.) -- (5.,5.) -- (5.,-5.);
  \draw [color=black] (-5.,5.) -- (-5.,-5.);
  \draw [color=black] (5.,5.) -- (5.,-5.);
  \draw [color=black] (-5.,5.) -- (5.,5.);
  \draw [color=black] (5.,-5.) -- (-5.,-5.);
 
 \draw [color=black] (10.8,-5.)-- (11.2,-5.);
  \draw [color=black] (10.8,5.)-- (11.2,5.);
  \draw [color=black] (10.8,2.5)-- (11.2,2.5);
    \draw [color=black] (10.8,0.)-- (11.2,0.);
      \draw [color=black] (10.8,1.25)-- (11.2,1.25);
       \draw [color=black] (10.8,1.25+1.25/2.)-- (11.2,1.25+1.25/2.);
       \draw [color=black] (10.8,1.25+1.25/2.+1.25/4.)-- (11.2,1.25+1.25/2.+1.25/4.);
   \draw [color=black] (11.,-5.) -- (11.,5.);

   \draw (13.,-3.) node[anchor=east] {\scriptsize{$\mathbf{T}_r^1$}};
   \draw (14.5,0.) node[anchor=east] {\scriptsize{$C(\mathbf{T}_r^1$)}};

 % \draw (5.,0.) node[anchor=west] {\scriptsize{$X^{-,r}$}};
 % \draw (-5.,0.) node[anchor=east] {\scriptsize{$X^{-,l}$}};
 
 % \draw (0.,5.) node[anchor=south] {\scriptsize{$X^{+,r}$}};
 % \draw (0.,-5.) node[anchor=north] {\scriptsize{$X^{+,l}$}};

  \draw[color=blue,<-] (3.,2.) -- (7.,3.);
  \draw[color=blue,<-] (3.,-2.) -- (7.,-3.);
 % \draw[color=blue,<-] (3.,-0.) -- (7.,-0.);

  \draw[color=blue,<-] (-3.,2.) -- (-7.,1.);
  \draw[color=blue,<-] (-3.,-2.) -- (-7.,-1.);
 % \draw[color=blue,<-] (-3.,-0.) -- (-7.,-0.);

  \draw[color=blue,->] (2.,-7.) -- (2.,-3.);
  \draw[color=blue,->] (-2.,-7.) -- (-3.,-3.);
%  \draw[color=blue,->] (-0.,-7.) -- (-0.,-3.);

  \draw[color=blue,->] (2.,7.) -- (3.,3.);
  \draw[color=blue,->] (-2.,7.) -- (-2.,3.);
 % \draw[color=blue,->] (-0.,7.) -- (-0.,3.);

  \draw[color=black,->] (-7.,-7.) -- (-5,-7.);
  \draw[color=black,->] (-7.,-7.) -- (-7,-5.);
  \draw (-7.,-6) node[anchor=east] {\scriptsize{$x^2$}};
  \draw (-6.,-7) node[anchor=north] {\scriptsize{$x^1$}};
  \draw (1.,-6) node[anchor=east] {\scriptsize{$\mathbf{T}_l^2$}};
  \draw (1.,6) node[anchor=east] {\scriptsize{$\mathbf{T}_r^2$}};
  \draw (-5.,0.) node[anchor=east] {\scriptsize{$\mathbf{T}_l^1$}};
  \draw (7.,0.) node[anchor=east] {\scriptsize{$\mathbf{T}_r^1$}};
\end{tikzpicture}
\caption{A schematic illustration of a trapping region $\mathbf{T}$, arrows indicate the possible directions of learning dynamics on the boundary (left), and a binary partitioning of the first right face of $\mathbf{T}$ (right).}\label{fig1}
\end{figure}

The visualization of assumption~\eqref{isolation} from Lemma~\ref{lem2} is presented in Figure~\ref{fig1}; the intuition behind it is that the learning operators $F_{ij}$ have to point inwards, into the trapping region, so their values have to be positive on left faces and negative on right faces. When the adaptive dynamics are not given explicitly (e.g. they depend on a reward from environment simulator), one may need to resort to verifying condition~\eqref{isolation} approximately, by evaluating the learning dynamics $F_{ij}$ on a finite subset of points, which provide good enough coverage of faces of $\mathbf{T}$.
If some analytical knowledge on learning dynamics is available,
we can verify~\eqref{isolation} rigorously (with sufficient numerical precision). 
For instance, assume that we know the upper bound for the Lipschitz constant of $F$ over $\mathbf{T}$, given by $L$.
Our verification is based on the following observation. We will check whether
\begin{equation} \pm F_{ij}(x)>0,\ x \in S,\end{equation}
where $S$ denotes either of the faces $\mathbf{T}^{ij}_l$, $\mathbf{T}^{ij}_r$, respectively, or their hyperrectangular subsets. Then it is enough to verify that either
\begin{equation}\label{ineq}
    \mp F_{ij}(C(S)) + L \operatorname{diam}(S)/2 < 0,
\end{equation}
where $C(S)$ is the baricenter (i.e. the centroid / intersection of diagonals) of $S$. Alternatively, we can show that 
\begin{equation}
    \pm F_{ij}(C(S))<=0,
\end{equation}
which will prove that the candidate $\mathbf{T}$ is not a trapping region.

If $S$ is the whole face of $\mathbf{T}$ (i.e. $\mathbf{T}_l^{ij}$ or $\mathbf{T}_r^{ij}$ for some $i,j$), then the verification of inequality~\eqref{ineq} can fail, even despite that $\mathbf{T}$ is a trapping region.
Therefore, we propose to adopt \emph{binary space partitioning} mechanism~\citep{fuchs1980visible} to iteratively subdivide faces of $\mathbf{T}$ into smaller hyperrectangles, until inequalities fail, or all hyperrectangles have been verified.  For details, we refer to the pseudocode in Algorithm~\ref{alg1}. The function SPLIT in Algorithm~\ref{alg1} splits a hyperrectangle $S$ into two disjoint non-empty hyperrectangles $S_1$, $S_2$, such that $S=S_1 \cup S_2$ in half, along the longest dimension of the hyperrectangle. 

\begin{theorem}
If $\mathbf{T}$ is a trapping region, Algorithm~\ref{alg1} is guaranteed to terminate in finite steps. Without loss of generality, assume that $\mathbf{T}$ is a unit hypercube. Then, the computational complexity of the algorithm is $O\left( log(L/2m^{*})^{\sum_{i=1}^{n} k_i -1 }{\sum_{i=1}^{n} k_i }\right)$, where 
\begin{equation}
m^{*}=\min_{i,j,x \in T^{ij}_{l,r} } |F_{ij}(x)|.\end{equation} Conversely, if Algorithm~\ref{alg1} terminates and returns true, $\mathbf{T}$ is a trapping region for learning rates as in Theorem~\ref{thm2}. 
\end{theorem}

\begin{algorithm}[t]
    \caption{Rigorous trapping region verification via binary space partitioning.}\label{alg1}
    \flushleft \textbf{Inputs:} Learning dynamics $F$, \\
        $\mathbf{T}=[x^{11}_-,x^{11}_+] \times \dots \times [x^{nk_n}_-,x^{nk_n}_+]$ -- a candidate for the trapping region,\\
    $L$ -- upper bound for Lipschitz constant of $F$ over $\mathbf{T}$. \\
    \textbf{Returns:} Is $\mathbf{T}$ a trapping region?
    
    \textbf{Start:}
    \begin{algorithmic}[1]
    \FOR{agent i in 1:$n$ in parallel}
    \FOR{coordinate j in 1:$k_n$ in parallel}
    \FOR{direction in \{left,right\} in parallel}
    \IF {direction is left}
    \STATE SETS\_TO\_CHECK = $\{ \mathbf{T}_l^{ij}\}$, $\delta = -1$
    \ELSE 
    \STATE SETS\_TO\_CHECK = $\{ \mathbf{T}_r^{ij}\}$, $\delta=1$
    \ENDIF 
    \WHILE{SETS\_TO\_CHECK $\neq \emptyset$}
    \STATE $S$ = SETS\_TO\_CHECK.POP()
    \STATE $C(S)=\operatorname{baricenter}(S)$
    \IF{$\delta F_{ij}(C(S)) \geq 0$}
    \RETURN \FALSE \ // no isolation
    \ELSIF{$\delta F_{ij}(C(S)) + L \operatorname{diam(S)}/2  \geq 0$} 
    \STATE // need subdivision to check isolation
    \STATE $S_1, S_2$=SPLIT($S$) // binary partitioning
    \STATE SETS\_TO\_CHECK.PUSH($S_1$, $S_2$)
    \ENDIF
    \ENDWHILE
    \ENDFOR
    \ENDFOR
    \ENDFOR
    \RETURN \TRUE
      \end{algorithmic}
\end{algorithm}

\subsection{Algorithmic verification: sampled learning dynamics}

In some situations, the exact learning dynamics are not available -- e.g. they depend on a reward, which can only be obtained from a real world or an experiment. Then, one has to resort to heuristic verification of trapping regions, by sampling points from the faces of the interval set. We provide the pseudocode for this situation in Algorithm~\ref{alg2}, and apply it in practice in traffic management example in Section~\ref{sec:traf}.

\begin{proposition}
Algorithm~\ref{alg2}  always terminates in finite steps, regardless of whether $\mathbf{T}$ is a trapping region or not.
The computational complexity of the algorithm is $O( 2M{\sum_{i=1}^{n} k_i} )$.
\end{proposition}

We remark that Algorithm~\ref{alg2} contains a naive, uniform sampling strategy, and one can envision a more sophisticated tree-like partitioning, like in Algorithm~\ref{alg1}, where we resample the regions in which we are closest to failing the isolation inequalities.
However, due to computational demands of our illustrative example, the traffic experiment in Section~\ref{sec:traf}, we have opted for uniform sampling, as it offers highest parallel execution potential,
and was most suited for execution in a computational cluster environment -- every sample evaluation can be executed as a separate process.

\begin{theorem}
Let $S^*$ be the set of all sampled points, $D$ be the size of mesh generated by the sample, i.e. 
\begin{equation}
D=\sup_{i,j,x \in T^{ij}_{l,r}} \min_{x^* \in S^*} ||x-x^*||.\end{equation} Also let \begin{equation}m^{*}=\min_{i,j,x^* \in S^*} ||F_{ij}(x^*)||\end{equation} quantify how close we were to fail verifying isolation over $S^{*}$. If Algorithm~\ref{alg2} returns true and $F$ is Lipschitz-continuous with Lipschitz constant $L<m^{*}/D$, then $\mathbf{T}$ is a trapping region for learning rates as in Theorem~\ref{thm2}.
\end{theorem}

\begin{algorithm}[t]
    \caption{Non-rigorous trapping region verification via sampling.}\label{alg2}
    \flushleft
    \textbf{Inputs:} Learning dynamics $F$,\\
    $F$ -- learning dynamics, can be only sampled (e.g. from simulator), \\
    $\mathbf{T}=[x^{11}_-,x^{11}_+] \times \dots \times [x^{nk_n}_-,x^{nk_n}_+]$ -- a candidate for the trapping region,\\
    $M$ -- sample size per face\\
    \textbf{Returns:} Is $\mathbf{T}$ a trapping region?
    
    \textbf{Start:}
    \begin{algorithmic}[1]
    \FOR{agent i in 1:$n$ in parallel}
    \FOR{coordinate j in 1:$k_n$ in parallel}
    \FOR{direction in \{left,right\} in parallel}
    \IF {direction is left}
    \STATE SET = $\mathbf{T}_l^{ij}$, $\delta = -1$
    \ELSE 
    \STATE SET= $ \mathbf{T}_r^{ij}$, $\delta=1$
    \ENDIF 
    // a uniformly spaced sample of $M$ points
    \STATE S = SAMPLE\_POINTS(SET, $M$) 
    \FOR{$x \in S$ in parallel}
    \STATE // $F$ evaluated on sample points
    \IF{$\delta F_{ij}(x) \geq 0$}
    \RETURN \FALSE \ // no isolation
    \ENDIF
    \ENDFOR
    \ENDFOR
    \ENDFOR
    \ENDFOR
    \RETURN \TRUE
      \end{algorithmic}
\end{algorithm}

\section{Examples}

In this Section we will provide examples of application of Algorithm~\ref{alg1} to  two systems with known dynamics -- a Generative Adversarial Network in Subsection~\ref{sec:gan} 
and of application of Algorithm~\ref{alg2} to a traffic learning system with dynamics provided by the system simulator in Subsection~\ref{sec:traf}. Additional example 
in a model of economic competition is provided in the Supplementary Material.

\subsection{Generative Adversarial Learning}\label{sec:gan}

Our first example is a toy system exemplifies the issue of non-convergence of multi-agent learning, but where trapping regions can be readily constructed.
Since the learning is non-convergent, methods based on Lyapunov functions, and regions of attractions of equilibria would not be applicable to this scenario.
We consider a parameterized family of learning systems,
where the parameter controls the coupling between learner rewards -- from completely decoupled, to strongly coupled. 
More concretely, agent one is in control of a continuous variable $\psi \in \mathbb{R}$, and agent two controls $\theta \in \mathbb{R}$. The rewards of each agent are the negative of the loss functions for each, and these are given by
\begin{equation}
 L_1(\psi, \theta) = \psi^4 +\epsilon \psi \theta ,
\end{equation}
and
\begin{equation}
 L_2(\psi, \theta) = \theta^4 - \epsilon \psi \theta ,
\end{equation}
for some positive, small $\epsilon$. 

Both agents use gradient descent on their respective loss functions, with a fixed step $\gamma$,
which leads to following update rules
\begin{align}\label{update}
    \begin{aligned}
        \psi_{t+1}&:=\psi_{t} - \gamma(4 \psi_{t}^3 + \epsilon \theta_{t}),\\
        \theta_{t+1}&:=\theta_t - \gamma(4 \theta_t^3 - \epsilon \psi_t),
    \end{aligned}
\end{align}
which, for short, we shall denote by $ (\psi_{t+1}, \theta_{t+1}) =: G(\psi_t,\theta_t)$.

Although a system with such prescribed loss functions is nothing more than a toy example, it serves to accentuate the problems of non-convergence, similar learning systems have been thoroughly analyzed in literature; this system in fact has the same update rules as the famously non-convergent Dirac-GAN example in~\citep{mescheder2018training} with the Wasserstein loss function, where both the generator and the discriminator apply an $L^4$ regularization term weighted by factor inversely proportional to $\epsilon$.

The dynamics of~\eqref{update} are surprisingly complicated for such low dimensional system.
The system possesses a single Nash equilibrium $(\psi,\theta)=(0,0)$ (also the only learning equilibrium), regardless of the value of $\epsilon$. For joint optimization, the equilibrium is always locally unstable (regardless of how small the system coupling parameter $\epsilon$ is),
and the learning trajectories starting from its near proximity diverge from it until they enter a cyclic regime. For initial conditions of larger norm, they converge towards the cyclic attractor, and never reach the equilibrium; in fact none of the other trajectories does. The divergence from the Nash equilibrium is formalized via the following proposition below (with $\lVert \cdot \rVert$ denoting $L^2$ norm):

\begin{proposition}\label{prop1}
 For any $\gamma>0$ and any $\epsilon>0$ there exist a value $R_0>0$, such that for any $(\psi_0,\theta_0)$ with $0 < \lVert (\psi_0,\theta_0) \rVert < R_0$ we have $\lVert G(\psi_0,\theta_0) \rVert > \lVert (\psi_0,\theta_0) \rVert.$ As a consequence, $(0,0)$ is a repelling equilibrium. 
\end{proposition}

\begin{figure}[t!]
\centering
\begin{overpic}[width=0.35\textwidth]{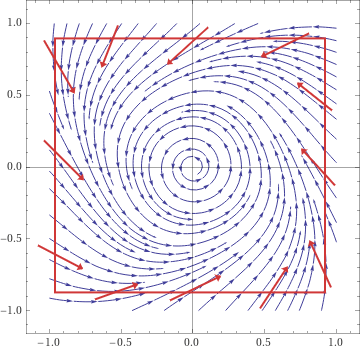}
\put(92,70){$\mathbf{T}$}
\put(96,0){$\theta$}
\put(3,95){$\psi$}
\end{overpic}
    \caption{Learning trajectories of regularized Dirac-GAN diverge from the center equilibrium, but can be proven to be contained by a trapping region. In such non-convergent scenario a construction of a Lyapunov function is impossible.}
    \label{fig:my_label}
\end{figure}

On the other hand,
it is easy to find trapping regions. We report that by executing Algorithm~\ref{alg1} we have successfully established existence of various trapping regions for different values of $\epsilon$, and computed $\gamma$ bounds via formula~\eqref{bound} (by making use of quantities obtained in the algorithm):
\begin{itemize}
    \item $\mathbf{T}=[-0.1,0.1]^4$ and $\epsilon \in \{0.01, 0.02, 0.03,0.04\}$ (we report failure, i.e. it is not a trapping region for $\epsilon=0.05$) and the upper bounds on $\gamma$ are given by 
    $\{4.6\times 10^{-3}, 7.9 \times 10^{-3}, 1.7 \times 10^{-3}, 10^{-17}\} $, respectively;
    \item 
    $\mathbf{T}=[-0.2,0.2]^4$ and $\epsilon \in \{0.05, 0.1, 0.15\}$
    (failure for $\epsilon=0.2$), and the upper bounds on $\gamma$ are given by $\{ 1.9\times 10^{-2}, 4.6 \times 10^{-4}, 2.0\times 10^{-4} \}$, respectively.
\end{itemize}
The Lipschitz constants in both examples were found analytically, by maximizing the $L^1$ norm of the total derivative $||D_{(\psi,\theta)}G||$ over $(\psi,\theta) \in \mathbf{T}$.
We remark that the closer we got to the point of failure, the more subdivisions were needed in the partitioning algorithm, however the execution time was near immediate -- a few seconds at most on a modern laptop, without leveraging parallelization. To contrast, a brute force optimization without verifying the trapping region would yield endless execution without convergence, and without any guarantees that learning will not diverge.

For this particular system, we can also prove the existence of an $\epsilon$-parameterized family of trapping regions theoretically, by the following proposition:

\begin{proposition}
 The square given by $[-\sqrt{\epsilon},\sqrt{\epsilon}]^2$ is a trapping region for step size $\gamma>0$ small enough. As a consequence, trajectories never leave $[-\sqrt{\epsilon},\sqrt{\epsilon}]^2$, and there is an equilibrium inside $[-\sqrt{\epsilon},\sqrt{\epsilon}]^2$ (it is in fact the global Nash equilibrium $(0,0)$). 
\end{proposition}

\subsection{Multi-Agent Traffic Management}\label{sec:traf}
\begin{figure}[t!]
\centering
    \includegraphics[width=0.38\textwidth, trim=530 200 800 200, clip]{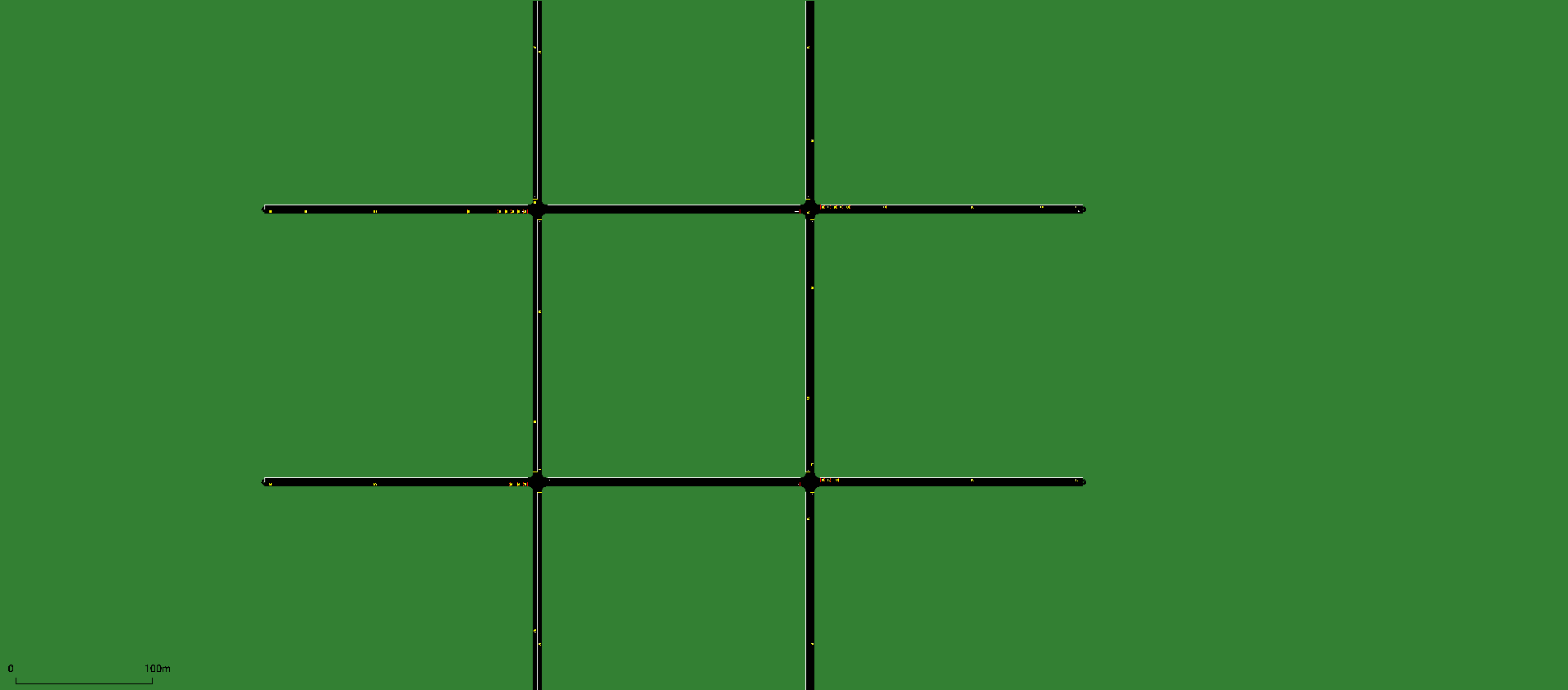}
    \caption{Multi-intersection traffic management problem map. Individual intersections
    optimize their own traffic demand by assigning green time to competing traffic streams.
    The safe set is defined by having minimally at least 20 seconds
    of uninterrupted green time for each traffic direction.}\label{fig:traf}
\end{figure}

Our second example is of a more practical nature.
We analyze a rectangular network of four signalized intersections,
each situated 200 meters from its two nearest neighbors,
as depicted in Figure~\ref{fig:traf}.
Each of the intersections controls traffic by alternating between one of two phases -- giving green to either the vertical or the horizontal stream of vehicles. The \emph{cycle time}, i.e. the total time for serving the horizontal and, subsequently, the vertical movement is set to 60 seconds.
For each episode of simulation, of length of two hours, each intersection can select a strategy from the continuous set $A=[0,60]$, which determines the amount of green seconds to be assigned to the first phase (\emph{the offset}). The remainder of the cycle is assigned to the second phase.
The vehicle streams are generated on all roads in all directions (i.e. east $\leftrightarrow$ west, and north 
$\leftrightarrow$ south),
and, for simplicity, we excluded left and right turning movements on intersections. The simulation is controlled by an open-source microscopic traffic simulator SUMO~\citep{lopez2018microscopic}, version 1.8.0. 
The episodical payoff for each intersection is the negative of the aggregate number of vehicle-seconds on all road lanes incoming to the intersection, therefore the goal of each intersection is to dispatch the incoming traffic as efficiently as possible. To ensure that the learning dynamics do not exhibit trivial symmetries, the simulations were performed for
a simple instance of asymmetrical demand. One vehicle would be spawned each ten seconds on the beginning of each of the outmost lanes of the network, with the exception of the northeast $\leftrightarrow$ northwest stream, where vehicles are spawned every five seconds. Analogous computations could have been performed for other traffic patterns.

For our experiment, the selection of the strategy by the learners is performed via decoupled gradient descent, as in Equations~\eqref{eq1} and~\eqref{eq2}. Each intersection controller estimates the gradient of own reward by difference quotients:
\begin{equation}
    \delta \nabla_{x_i} R_i(x) \approx R_i(x_i, x_{-i}) - R_i(x_i+\delta, x_{-i})
\end{equation}
for some small $\delta$ (in our experiments $\delta=0.1$). The adaptation rate $\gamma$ is set to $10^{-6}$. Such settings were chosen as they would give satisfactory results for learning on one intersection, while keeping other intersections fixed.

\begin{figure}[t!]
\centering
\begin{overpic}[width=0.41\textwidth]{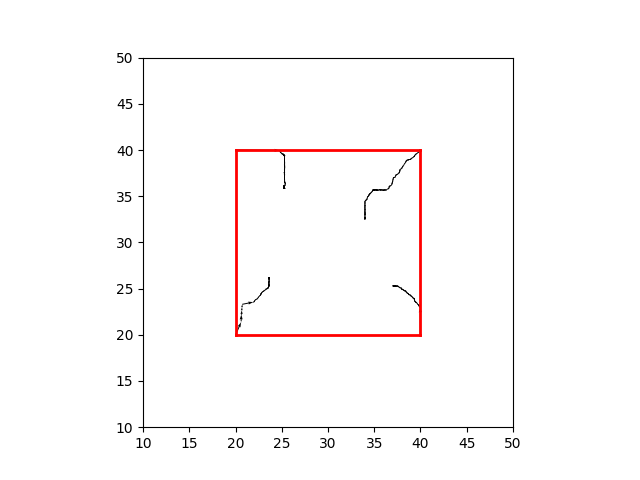}
\put(67,40){$\mathbf{T}$}
\put(72,10){\tiny{$NW$}}
\put(23,62){\tiny{$NE$}}
\end{overpic}
    \caption{The trapping region $\mathbf{T}$ (in red) and learning trajectories in the traffic control scenario (in black) -- a projection onto first two dimensions of joint strategy space (corresponding to offsets of northeastern and northwestern intersections). Learning curves, which begin on the boundary of the trapping region do not escape it, and evolve in the interior of the set. }
    \label{fig3}
\end{figure}

As discussed previously, non-convergence is an undesirable learning effect in such traffic management scenario, as one would like to ensure that learning always stays within some predetermined bounds, so minimal green time can be given to vehicle flows and pedestrians within each cycle.
As a reasonable prerequisite we assume that each phase should be given at least 20 seconds of green time,
which translates to the candidate for a trapping region given by $\mathbf{T}= [20,40]^4$.
From the nature of the problem, we expect the reward function to be continuous, but we do not have an analytical formula for it. Therefore, we apply Algorithm~\ref{alg2} and sample faces of $\mathbf{T}$ with a uniform rectangular grid of five points in each direction ($M=125$). This part of computation is parallelized over multiple threads, and so it does not significantly increase verification time.

The numerical evaluation of isolation inequalities~\eqref{isolation} is successful; and, according to Theorem~\ref{thm1} we conclude that each learning trajectory that starts in $\mathbf{T}$ stays in $\mathbf{T}$, regardless of whether it converges, and that there is a learning equilibrium in $\operatorname{int} \mathbf{T}$.
We illustrate this by plotting projections of the trapping region and four learning trajectories in Figure~\ref{fig3}.
We remark that due to parallelization, verifying the region was much more efficient computationally, than computing learning trajectories.
It took about half an hour on 32 CPUs,
whereas computation of depicted 500 steps of each learning trajectory took about eight hours, and could utilize only one CPU per trajectory.
The experiment used AMD 7452 and AMD
7502P CPUs, 2.35 and 2.5 Ghz respectively.
It would have been technically possible to execute the algorithm on a GPU, however in this example it would bring no advantage, as the most time consuming part was obtaining reward difference quotients from the system simulator, which can only run on CPUs.

\section{Applicability, limitations and future work}

In this paper we have introduced the method of trapping regions,
which can be used to circumvent safety problems caused by non-convergence in multi-agent learning.
We have presented algorithms for verification of trapping regions, and theoretical results on the implications for safety and provided examples in GAN learning, in an applied traffic management scenario,
and (in the Supplementary Material) in a standard mathematical model of economic competition.

Our examples are relatively low-dimensional.  
We however remark, that even low-dimensional learning can be non-convergent and highly unpredictable, and therefore pose safety concerns (e.g. the chaotic example of chaos in a simple two player game~\cite{Sato2}). Moreover,
low-dimensionality of strategy spaces does not mean that the
learning systems need to be trivial; for instance agents can be
controlled by high-dimensional pre-trained neural networks
with the last layer being retrained online. Our algorithms are well suited to deal with such scenarios. The extension of the method to high-dimensional settings is a challenge for future research, due to exponential complexity of verification algorithms w.r.to the joint action space.
We see possibilities for exploiting symmetries of the action space as a method for dimensionality reduction: e.g., in mean field games where an infinite amount of identical agents share same learning dynamics~\cite{yang2018mean},
or by employing coordination graphs~\citep{kuyer2008multiagent}.

\section*{Acknowledgements}
This project was supported by the European Research Council (ERC) 
under the European Union's
Horizon 2020 research 
and innovation programme (grant agreement No.~758824 \textemdash INFLUENCE).
 \begin{figure}[h!]
   \centering
%\hspace{-.25in}
\includegraphics[width=0.4\columnwidth]{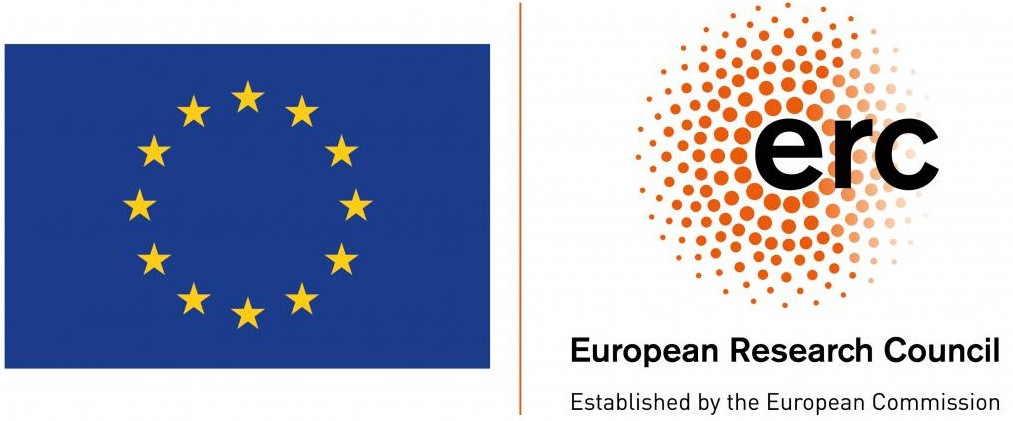}
\end{figure}

\bibliographystyle{named}
\bibliography{bibliography}

\section*{Supplementary Material}

\textbf{Proof of Theorem 1}
\begin{proof} 
The first part of the assertion follows directly from the definition.
The second part is a consequence of the Brouwer fixed point theorem --
every continuous self-map of a compact, convex subset of $\mathbb{R}^N$ has a fixed point. 
\end{proof}

\noindent \textbf{Proof of Lemma 1}
\begin{proof}
Assume otherwise; then there must exist a trajectory $\{x_t\}_t$, which leaves $\mathbf{T}$. Let $t_0$ be the first time for which $x_{t_0} \notin \mathbf{T}$. 
Let $\epsilon$ be small enough, such that if $\operatorname{dist}(x,\partial \mathbf{T}) < \epsilon$, then $x+\gamma F(x) \in \operatorname{int} \mathbf{T}$. Such $\epsilon$ can be found, as $F$ is continuous and $\mathbf{\partial{T}}$ is compact. 
For $\gamma$ suitably small we have $\operatorname{dist}(x_{t_0-1}, \partial \mathbf{T}) < \epsilon$, which leads to a contradiction with $x_{t_0} \notin \mathbf{T}$.
\end{proof}

\noindent \textbf{Proof of Theorem 2}
\begin{proof}
The proof goes along the same lines as the proof of Lemma~1. Since $F$ is Lipschitz-continuous, for all $x=\{x^{ij}\} \in \mathbf{T}$ such that 
\begin{equation}\label{helper}
\max_{ij} |x^{ij} - x^{ij}_{-}| < \frac{\min_{y \in \mathbf{T}_l^{ij}} F_{ij}(y)}{L}
\end{equation}
we have $F_{ij}(x) > 0$. We would like to ensure that no trajectory that starts from a point $x \in \mathbf{T}$ is to leave $\mathbf{T}$ by having too small value of one coordinate, and to this end, for any given pair of indices $i,j$ we consider two cases. 

Firstly, if $F_{ij}(x) \geq 0$, then \begin{equation}x^{ij}+\gamma F_{ij}(x) \geq x^{ij}\end{equation} for all $\gamma>0$ and the inequality is strong if $x \in \mathbf{T}_l^{ij}.$
Secondly, if $F_{ij}(x) < 0$,
then it is enough to show that \begin{equation}x^{ij} + \gamma F_{ij}(x) \geq x^{ij}_{-}.\end{equation}
From~\eqref{helper} we know that $|x^{ij}_{-} - x^{ij}| >\max_{y\in \mathbf{T}^{ij}_l} F_{ij}(y)/L$.
Therefore, it is enough that \begin{equation}\gamma |F_{ij}(x)| < \max_{y\in \mathbf{T}^{ij}_l} F_{ij}(y)/L.\end{equation}
By a mirror argument for faces $\mathbf{T}_{r}^{ij}$, and taking a minimum over all faces, we conclude that inequality~(8) is a sufficient condition for $\mathbf{T}$ to be a trapping region.
\end{proof}

\noindent \textbf{Proof of Theorem 3}

\begin{proof}

Algorithm~1 contains several \texttt{for} loops, which always end in finite time, and one $\texttt{while}$ loop (line 8), termination of which being not immediately clear. We will show that for any given face $\textbf{T}_l^{ij}$ or $\textbf{T}_r^{ij}$ this loop indeed terminates. To fix our attention, let us assume we have a face of form $\textbf{T}_l^{ij}$ for some $i,j$ (analogous analysis can be performed for $\textbf{T}_r^{ij}$, with adjusted signs). There are three possible scenarios. If $F_{ij}( x ) > 0\ \forall x \in \mathbf{T}_l^{ij}$ (when $\mathbf{T}$ is a trapping region), then
    (with enough numerical precision) the evaluation of inequalities
    \begin{equation}
        -F_{ij}(C(S)) > 0
    \end{equation}
    for $S \subset \mathbf{T}_l^{ij}$ will always return \texttt{False},
    and the loop will always progress to the \texttt{else if} line.
    Since $\textbf{T}_l^{ij}$ is compact, there will always be an $\epsilon>0$ such that $$F_{ij}( x ) > \epsilon\quad \forall x \in \textbf{T}_l^{ij}.$$
    Moreover, after a finite amount of operations, the diameter of any set remaining $S \in \text{SETS\_TO\_CHECK}$ will become smaller than $\epsilon/L$ (as it has been either subdivided enough times, or removed). At this point, the evaluation of the \texttt{else if} condition will also always fail, with enough numerical precision,
    and the loop will pop all the remaining sets in $\text{SETS\_TO\_CHECK}$, and eventually (after analogous calculations are performed for other faces), return true. By Lemma 2, these evaluations prove that $\mathbf{T}$ is indeed a trapping region. 
    
    The computational complexity can be argued as follows. Let $D$ be the supremum of diameters of sets in the inner loop of the algorithm. In the worst case, from Lipschitz continuity, we will need $D/2.<m^*/L$ to verify the isolation inequalities. This will happen after approximately $\operatorname{log} (L/2m^*)$ subdivisions in each dimension of each face. By taking the product over $\sum_{i=1}^{n} k_i-1$ dimensions of the hyperface, and a  sum over all $2\sum_{i=1}^{n} k_i $ faces of the boundary of $\mathbf{T}$, we arrive at the desired formula.  
    
    In the second scenario, there is an $x \in \textbf{T}_l^{ij}$ such that $F_{ij}(x) = 0$, but
    $F_{ij}(y)>0$ for $y \in \textbf{T}_l^{ij}$ sufficiently close to $x$.  
    This situation is degenerate, and corresponds to an $\emph{internal tangency}$ of the learning direction with the face of set $\mathbf{T}$.
    In such situation, the \texttt{while} loop of the algorithm might never finish, as the point $x$ might never be directly evaluated (e.g. it could have irrational coordinates). If an internal tangency occurs, $\mathbf{T}$ is not a trapping region.
    
    In the third situation, there is no internal tangency, and in addition there exists an $x \in \textbf{T}_l^{ij}$ such that $F_{ij}(x) < 0$.
    That means there exists a set $R$ of non-empty interior such that $x\in R$ and 
    \begin{equation}
        F_{ij}(y) < 0\quad \forall y \in R.
    \end{equation}
    After enough subdivisions the diameters of sets will be so small that we will have a set $S \in \text{SETS\_TO\_CHECK}$ such that $S \subset R$. Then $F_{ij}(C(S))<0$ and the $\texttt{while}$ loop will exit at the $\texttt{return false}$ statement.
\end{proof}

\noindent \textbf{Proof of Proposition 1}
\begin{proof} 
Termination in finite steps follows from that the algorithm contains \texttt{for} loops only, with a finite number of $O(M)$ instructions, all completing in finite time, within each loop. 

\end{proof}

\noindent \textbf{Proof of Theorem 4}

\begin{proof}
Let $x \in \textbf{T}_{l}^{ij}$, and let $x^*$ be the closest point in $S^* \cap \textbf{T}_{l}^{ij}$
From Lipschitz continuity we have 
\begin{equation}
||F_{ij}(x) - F_{ij}(x^*)|| \leq DL \leq m^*,
\end{equation}
which, coupled with $||F_{ij}(x^*)|| \geq m^*$, implies $F_{ij}(x)>0$. A mirror argument follows for $x \in \textbf{T}_{r}^{ij}$.

\end{proof}
\pagebreak

\noindent \textbf{Proof of Proposition 2}
\begin{proof}
Let us fix $\gamma$ and $\epsilon$.
We have:
\begin{align}
\begin{aligned}
    &\lVert G(\psi_0,\theta_0) 
    \rVert^2
    \\&=
    (1+\epsilon\gamma)\psi_0^2 + (1+\epsilon\gamma)\theta_0^2
    + \text{h.o.t} \\&> \lVert (\psi_0,\theta_0) 
    \rVert^2
    \end{aligned}
\end{align}
for $\psi_0,\theta_0$ small enough.
In the above, h.o.t. are higher order terms in $\psi,\theta$, i.e. polynomials of degree 2 or higher. 
\end{proof}

\noindent \textbf{Proof of Proposition 3}
\begin{proof}
It is enough to verify the isolation conditions given by Lemma~\ref{lem2}. 
The dynamics of the learning system~\eqref{update} can be expressed
in the standard form~\eqref{eq1}, by setting
\begin{align*}
\begin{aligned}
    F_1(\psi, \theta)&:=-4 \psi^3 - \epsilon \theta,\\
    F_2(\psi, \theta)&:= - 4 \theta^3 + \epsilon \psi.
\end{aligned}
\end{align*}
\noindent
We first observe that $F_1$ is monotonically decreasing in $\theta$ and $F_2$ is monotonically increasing in $\psi$.
Therefore, to verify the isolation inequalities (7) it is enough to verify them on the corner points of $\mathbf{T}=[-\sqrt{\epsilon}, \sqrt{\epsilon}]^2$.
Indeed, we have
\begin{align*}
    F_1(-\sqrt{\epsilon}, \sqrt{\epsilon})&=3\epsilon \sqrt{\epsilon}>0,\\
    F_1(\sqrt{\epsilon}, -\sqrt{\epsilon})&=-3\epsilon \sqrt{\epsilon}<0,\\
    F_2(-\sqrt{\epsilon}, -\sqrt{\epsilon})&=3\epsilon \sqrt{\epsilon}>0,\\
    F_2(\sqrt{\epsilon}, \sqrt{\epsilon})&=-3\epsilon \sqrt{\epsilon}<0.
\end{align*}
\noindent
Furthermore, we have $\operatorname{argmin}_{\psi} L_1(\psi, 0) = 0$
and $\operatorname{argmin}_{\theta} L_2(0, \theta) = 0$, so the joint strategy $(0,0)$ is a Nash equilibrium.
\end{proof}

\noindent \textbf{The Cournot Duopoly example}

We consider gradient learning in the Cournot oligopoly model of economic competition~(Cournot, 1838).
In the classical, convex, setting it is a simple system, which admits explicit attracting equilibriated strategy profiles.
We use it as another example of application of Algorithm~1, with more practical value (due to economic interpretation) than the GAN, although with simpler dynamics.

Cournot oligopoly is a continuous action game,
where players represent companies, which try to maximize their profits earned from production and supply of commodities. 
The
$i$-th company controls the level of output of their product, denoted by a positive real variable $x^i$, with $x=\{x^i\}_i$. The production cost of the good is represented by a continuous real function
$c_i(x^i)$, and the revenue from selling the commodity
is given by a product of the output $x^i$ and of the price $d_i(x)$, the latter being itself a function
of production outputs. In the standard setting, the cost function is linear, i.e.
$    c_i(x^i) = c_ix^i,$
and the price function is an affine function of the form:
\begin{equation}
    d_i(x^1, \dots , x^n) = a-b_{ii} x^i - \sum_{j \neq i} b_{ij} x^j, 
\end{equation}
with $c_i, b_{ij} \in \mathbb{R}^+ \cup \{0\}$. 
Finally, the payoff for player $i$ is the difference between its revenue and cost:
\begin{equation}
    u_i(x) = x^i d_i(x^1, \dots, x^n) - c_i(x^i).
\end{equation}
The rationale behind the formula 
for the price function, is that the trading price of $i$-th commodity 
is negatively correlated with its own supply, 
and is to some extent replaceable by commodities produced by other companies -- which is quantified by the magnitude of coefficients $b_{ij}, i \neq j$.
Therefore, one usually assumes that $b_{ii} \gg b_{ij}$, $j \neq i$.
Furthermore, by suitable rescaling, without loss of generality we can set $a:=1$.

We consider a duopoly model and use individual gradient ascent as the update rule. The gradients are given by
\begin{align}
\begin{aligned}
    \nabla u_1(x) &=d_1(x)-x^1-c_1 ,\\
    \nabla u_2(x) &= d_2(x)-x^2-c_2.
\end{aligned}
\end{align}

In numerical experiments it was relatively easy to construct trapping regions for various combinations of parameters, as long as the coupling (indicated by parameters $b_{12}$ and $b_{21}$) was low enough.
For instance, we report that we have verified a trapping region of the form
\begin{equation}
    \mathbf{T} = [0.15, 0.3] \times [0.1, 0.3], 
\end{equation}
via Algorithm~1, for the following parameters:
\begin{align}
\begin{aligned}
    b_{11}=b_{22} &= 1,\\
    b_{12} &=0.2, \\
    b_{21} &= 0.1,\\
    c_1 = c_2 &= 0.5,
\end{aligned}
\end{align}
with upper bound on $\gamma$ given by $2.5\times 10^{-3}$.
The upper bound for the Lipschitz constant was found by computing the $L^1$ norm of the learning operator 
\begin{equation}
    ||DF||_{L_1}=4+b_{12}+b_{21}.
\end{equation}
By Theorem~1, we deduce that all learning trajectories that 
begin in $\mathbf{T}$ do not ever leave $\mathbf{T}$ -- leading to a practical conclusion,
that for any of such starting conditions the production of neither of companies would never drop to zero.

\end{document}